\documentclass[lettersize,journal]{IEEEtran}
\pdfoutput=1
\usepackage{amsmath,amsfonts}
\usepackage{amssymb}
\usepackage{mathtools}
\usepackage{amsthm}
\usepackage{algorithmic}
\usepackage{algorithm}
\usepackage{array}
\usepackage{booktabs}
\usepackage[caption=false,font=normalsize,labelfont=sf,textfont=sf]{subfig}
\usepackage{textcomp}
\usepackage{stfloats}
\usepackage{url}
\usepackage{verbatim}
\usepackage{graphicx}
\usepackage{cite}
\usepackage{nicefrac}       
\usepackage{tikz}
\usetikzlibrary{shapes.geometric}
\usepackage[showseconds=false,showzone=false]{datetime2}

\usepackage{mathtools}
\DeclarePairedDelimiterX{\norm}[1]{\lVert}{\rVert}{#1}

\usepackage{array}
\usepackage{multirow}

\def\y{{\mathbf y}}
\def\x{{\mathbf x}}
\def\f{{\mathbf f}}
\def\g{{\mathbf g}}
\def\u{{\mathbf u}}
\def\v{{\mathbf v}}
\def\L{{\cal L}}
\def\G{{\cal G}}
\def\fone{{\mathbf{f}^{(1)}}}
\def\ftwo{{\mathbf{f}^{(2)}}}

\def\fk{{\mathbf{f}^{(k)}}}

\allowdisplaybreaks

\begin{document}

\title{Timestamped Embedding-Matching\\ Acoustic-to-Word CTC ASR}

\author{Woojay Jeon
\thanks{Woojay Jeon is with Apple in Cupertino, California, U.S.A. (e-mail: woojay@apple.com)}}


\markboth{Manuscript Preprint}%
{Shell \MakeLowercase{\textit{et al.}}: Submission Pre-Print}


\maketitle

\begin{abstract}
In this work, we describe a novel method of training an embedding-matching word-level connectionist temporal classification (CTC) automatic speech recognizer (ASR) such that it directly produces word start times and durations, required by many real-world applications, in addition to the transcription.
The word timestamps enable the ASR to output word segmentations and word confusion networks without relying on a secondary model or forced alignment process when testing.
Our proposed system has similar word segmentation accuracy as a hybrid DNN-HMM (Deep Neural Network-Hidden Markov Model) system, with less than 3ms difference in mean absolute error in word start times on TIMIT data.
At the same time, we observed less than 5\% relative increase in the word error rate compared to the non-timestamped system when using the same audio training data and nearly identical model size.
We also contribute more rigorous analysis of multiple-hypothesis embedding-matching ASR in general.
\end{abstract}

\begin{IEEEkeywords}
Automatic speech recognition, CTC, embedding matching.
\end{IEEEkeywords}

\section{Introduction}

\IEEEPARstart{R}{ecently,} embedding-matching acoustic-to-word (A2W) ASR \cite{settle-2019,yen-2023} has gained interest due to its flexibility in accommodating out-of-vocabulary (OOV) words.
The entire vocabulary of the ASR is explicitly represented by a matrix of acoustic word embeddings \cite{jeon-ane-2020,he-2017} generated by an external word encoder.
The matrix can be treated independently of the rest of the system, and can be extended with more columns (words) with no systemic limits.
This modularity is attractive for scenarios where the ASR's vocabulary must be extended with dynamic out-of-vocabulary (OOV) words -- especially user-dependent named entities -- on-the-fly for every utterance.
Named entities can include contact names, on-device song titles, and location-dependent points of interest, all of which can change over time for a given user, and vastly differ from user to user.
The system is also scalable because it distills continuous word recognition into a series of nearest neighbor searches over a vector space, which is highly parallelizable in modern hardware \cite{garcia-2010}.


In this paper, a \emph{timestamped} embedding-matching A2W system is proposed, where the underlying connectionist temporal classification (CTC) model is trained such that for every hypothesized word, the model also estimates a word timestamp -- the start time and duration of the word.
The model learns this capability from ground truth segmentations provided during training.
The timestamps allow the decoder to directly produce the word segmentations of recognition results, as well as word confusion networks \cite{mangu-2000, jeon-conf-2020}, which are vital to many speech-based applications.
Audio segmentations are used for keyword search \cite{gui-xin-shi-2021}, as well as the ``intercom'' feature of digital assistants.
Word confusion networks are needed to rank and display word alternatives in a mobile user interface \cite{larson-2003}, semantically parse the utterance \cite{tur-2013}, and mitigate errors in spoken language translation \cite{ng-2015}.

A ``pure'' end-to-end ASR where a static monolithic model produces only the word transcription is of limited use in practical applications \cite{jinyu-li-2022}.
The ability to dynamically modify or augment the vocabulary, as well as produce accurate word segmentations is also needed.
The timings naturally produced by conventional CTC models are generally too inaccurate \cite{yang-2022} to be relied on.
Furthermore, the interval between word onsets are poor estimates of word duration when there is significant silence between the words.
A common remedy is to run a second-pass forced alignment using a hybrid DNN-HMM acoustic model \cite{mohamed-2012} to obtain the segmentation, but this is inefficient, and not scalable for multiple hypotheses.
Other works used a constrained attention head in a second-pass rescoring model to emit timings from an RNN transducer model \cite{sainath-2020}, combined a CTC model with a separate framewise phoneme classifier \cite{yang-2021}, or used a separate timestamp estimator model with a CTC-attention model \cite{yang-2022}.
Slightly related to the timestamp problem is the streaming latency problem, where E2E models are trained to reduce the latency of word outputs \cite{shinohara-2022}.
Such methods cannot be a substitute for explicit timestamp estimation, since they are essentially dealing with word \emph{end} times, not start times and durations, and furthermore it is likely impossible to achieve zero streaming latency.

In our proposed system, the word-level CTC model directly provides a timestamp with each word, with no additional model or alignment process at runtime.
A DNN-HMM ASR is needed to force-align the audio to the transcriptions to obtain ground truth segmentations used to train the CTC model, but this force-alignment only needs to be done once at \emph{data collection time}, meaning that the ground truth can be permanently stored and reused for the lifetime of the training data.
To the best of our knowledge, no CTC model with such timestamping capability exists in the literature.
Our system also produces timestamps for every hypothesized word, not just the 1-best words.
If a CTC model can directly produce timestamps without requiring an additional system at run-time, the system becomes simpler and faster while consuming less power and computational resources.

We will begin with some theoretical analysis that was absent in our previous study \cite{yen-2023} on embedding-matching word CTC ASR.
In particular, the multiple-hypothesis system we proposed in \cite{yen-2023} left open the question of why a simple summation of scores would not result in the multiple embeddings collapsing into the same embedding.
In this paper, we provide some theoretical developments that address this question.
We will then describe and analyze the proposed timestamped system, followed by experimental results on its word segmentation accuracy and its impact on recognition accuracy.

\section{A More Rigorous Analysis of Embedding-Matching Word CTC}
\label{sec:analysis1}

\subsection{Review of Conventional Word CTC}
 
We begin with a brief review of conventional word CTC ASR. The word CTC loss \cite{graves-2006, soltau-2017} is the negative sum of the log posterior over all training samples $\mathcal{D}$ where each sample consists of an audio input $X$ and a reference word sequence $W$:
\begin{equation}\label{eq:ctc-loss}
\L = - \hspace{-0.5em} \sum_{(X, W) \in \mathcal{D}} \hspace{-0.5em} \ln \hat{P}(W|X).
\end{equation}
Each $\hat{P}(W|X)$ is the sum of the individual posteriors of all possible word sequences $\pi$ in a set $\mathcal{B}^{-1}(W)$ that map to $W$, where the ``words'' in $\pi$ include a ``blank'' word:
\begin{equation}
\hat{P}(W|X) = \hspace{-0.5em} \sum_{\pi \in \mathcal{B}^{-1}(W)} \hspace{-0.5em} \hat{P}(\pi | X).
\end{equation}
Each sequence posterior $\hat{P}(\pi|X)$ is in turn the product of the frame-wise scores for every word in the sequence:
\begin{equation}
\hat{P}(\pi|X) = \prod_{t=1}^T p_{t,\pi_t},
\end{equation}
where $T$ is the number of frames and the score $p_{t,i}$ for every word $w_i$ in a vocabulary of $n$ words at time $t$ is obtained by applying a softmax to the output of a neural network:
\begin{equation}\label{eq:softmax}
p_{t,i} = \frac{\exp(s_{t,i})}{\sum_{j=1}^{n} \exp(s_{t,j})}, \ 1 \le t \le T, \ 1 \le i \le n 
\end{equation}

\subsection{Analysis of Single-Hypothesis System}
\label{subsec:single-embedding}

An ``embedding-matching'' word CTC \cite{yen-2023, settle-2019} differs from conventional word CTC in how each $s_{t,i}$ in Equation \eqref{eq:softmax} is computed.
Instead of training a neural network to freely and directly produce $s_{t,i}$, we train the network to produce an audio embedding vector $\f_t$.
Each $s_{t,i}$ is explicitly constrained to be the negative $\L_2$ distance as in our previous study \cite{yen-2023} (or inner product \cite{settle-2019}) between $\f_t$ and a text embedding vector $\g_i$ precomputed by an external acoustic text encoder \cite{jeon-ane-2020,he-2017} for word $w_i$.
Only the blank word's score $s_{t,0}$ is produced directly:
\begin{equation} \label{eq:score}
s_{t,i} = 
\begin{cases}
\displaystyle -b_{t}^2 & (i=0) \\[1ex]
\displaystyle -\Vert\f_t - \g_i \Vert^2 & (i=1, \cdots, n)
\end{cases}
\end{equation}
In practice, as we noted previously \cite{yen-2023}, all $\g_i$ are stored as the columns of matrix $G$, and a sequence of matrix operations result in a vector $\mathbf{s}_t$ containing the $s_{t,i}$'s.

Although the \emph{pre-softmax} score $s_{t,i}$ is maximized when $\f_t = \g_i$, the \emph{post-softmax} score $p_{t,i}$ is not necessarily maximized, due to the effect of the competing scores in the denominator in \eqref{eq:softmax}.
This would imply that the CTC criterion -- which essentially attempts to maximize $p_{t,i}$ for reference word $i$ at some time $t$ -- may not \emph{strictly} try to make $\f_t$ match $\g_i$.
For example, consider a vocabulary of 3 words with 1-dimensional embedding vectors $\g_1=[1.0]$, $\g_2=[2.0]$, and $\g_3=[4.0]$. $s_{t,2}$ is maximized when $\f_t=[2.0]$, but $p_{t,2}$ is maximized when $\f_t=[2.385]$.

We can argue, however, that if the embedding dimensions are sufficiently higher than 1, we can expect $\f_t \approx \g_i$ when $p_{t,i}$ is maximized.
The log of the softmax in Equation \eqref{eq:softmax} is
\begin{equation}\label{eq:log_softmax}
\ln p_{t,i} = - \ln \bigg[ 1 + \sum_{j \ne i}^{n} \exp(s_{t,j} - s_{t,i}) \bigg].
\end{equation}
A well-known interpretation of the softmax \cite{GoodBengCour16} is that if $s_{t,i}$ is much higher than all other $s_{t,j}$, then the summation term above will be close to 0, resulting in $p_{t,i} \approx 1$ while $p_{t,j} \approx 0 \ (j \ne i)$. 
In our case, if $\f_t=\g_i$, we will have $s_{t,i}=0$, while $s_{t,j}$ will be strongly negative if $\g_i$ is sufficiently far away from the rest of the vectors.
This effect is more apparent for higher embedding dimensions, since more negative terms are added to $s_{t,j}$.
Hence, we conclude that $\f_t=\g_i$ asymptotically maximizes $p_{t,i}$.

Other studies \cite{settle-2019,collobert-2020} used an inner product instead of negative Euclidean distance for $s_{t,i}$. In Appendix \ref{appendix-a}, we show that a similar argument as above can be made for the inner product that for sufficiently high dimensions, one can expect that $\f_t=\g_i$ will asymptotically maximize $p_{t,i}$.

\subsection{Analysis of Multiple-Hypothesis System}
\label{subsec:multiple-embedding}

\begin{figure}
     \centering
     \includegraphics[width=1.05\linewidth]{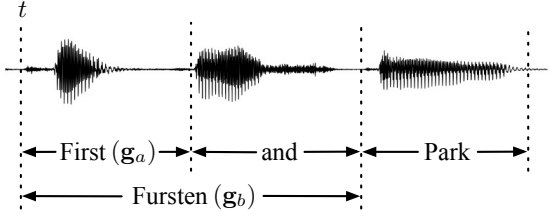}
     \caption{
     Example where two training samples have nearly-identical audio but different word transcriptions ``First and Park'' and ``Fursten Park.''
     The multiple-hypothesis embedding-matching system is induced to produce, for the same input audio, two different embeddings $\fone=\g_a$ and $\ftwo=\g_b$ at some point in time $t$ (which may be anywhere, not necessarily at the beginning of the words as shown in the diagram), where $\g_a$ is the text embedding for ``First'' and $\g_b$ is the text embedding for ``Fursten.''
     }
     \label{fig:fursten-wav1}
\end{figure}

In the multiple-embeddings approach we proposed in \cite{yen-2023}, the underlying model produces $L$ embeddings $\f_t^{(1)}, \cdots, \f_t^{(k)}$, and their pre-softmax scores are summed:
\begin{equation}\label{eq:logical-or}
    s_{t,i} = 
    \begin{cases}
    \displaystyle -b_{t}^2 & (i=0) \\[1ex]
    -\sum_{k=1}^L \Vert\f_t^{(k)} - \g_i \Vert^2 & (i=1, \cdots, n)
    \end{cases}
\end{equation}
The training can be best understood with a simple example, with $L=2$.
Consider a training dataset of just two utterances that have almost the same audio but different transcriptions: ``Directions to First and Park'' and ``Directions to Fursten Park.''
We can further simplify this as the \emph{same} audio producing two different outputs, as shown in Fig. \ref{fig:fursten-wav1}.

Assuming the language model does not contain the compound word ``first\_and'', all three words ``First,'' ``and,'' and ``Fursten'' must be separately hypothesized by the acoustic model.
Furthermore, to avoid a result like ``First Fursten'' or ``Fursten First,'' the system will ideally output ``First'' and ``Fursten'' simultaneously (Figure \ref{fig:decode-baseline} shows an example). 

Suppose $w_a=$``First'' and $w_b=$ ``Fursten'', and their text embedding vectors are $\g_a$ and $\g_b$, respectively.
The intent of the multiple-hypothesis system is to allow the network, at some point in time $t$, to output $\f_t^{(1)} = \g_a$ and $\f_t^{(2)}=\g_b$ (or vice versa) for the input audio in Fig. \ref{fig:fursten-wav1}.

Since the training criterion is to maximize the posterior for both training samples, we can consider the post-softmax score $p_a$ and $p_b$ at time $t$. 
We need not consider the case where $p_a$ and $p_b$ are maximized at different times because that would result in the model outputting the two words in sequence, i.e., ``First Fursten'' or ``Fursten First.''
Since we are dealing with only one specific value of $t$, we drop $t$ from our notation for simplicity.
The CTC training will effectively try to maximize
\begin{equation}\label{eq:joint-posterior}
     \ln(p)= \ln(p_a p_b) = s_a + s_b - 2 \ln \sum_j \exp(s_j).
\end{equation}
We make a similar argument as in Sec. \ref{subsec:single-embedding} that the maximum $s_c$ among all $s_j$ is distinct enough so that
\begin{equation}\label{eq:approx-max}
     s_c \triangleq \max_j{s_j} \approx \ln \sum_j \exp (s_j).
\end{equation}
If we assume for a moment that $\g_j$ can be any value,
\begin{equation}
     \frac{\partial s_j}{\partial \g_j} = 2(\fone - \g_j) + 2(\ftwo-\g_j)
\end{equation}
and $\partial^2 s_j / \partial \g_j^2 = -4$ (elementwise),
which means $s_j$ is maximized by $\g_j = \frac{1}{2}(\fone + \ftwo)$.
In reality, $\g_j$ cannot be arbitrary, and can only be chosen from our fixed vocabulary.
But if our vocabulary is large, we can assume that such a $\g_j$ exists.
Since $s_c$ is the maximum $s_j$ in \eqref{eq:approx-max}, we have $\g_c = \frac{1}{2}(\fone + \ftwo)$.
Next, we differentiate \eqref{eq:joint-posterior} to obtain
\begin{equation}
     \frac{\partial \ln(p)}{\partial \fone} = 2(\g_a + \g_b) - 2(\fone + \ftwo)
\end{equation}
and $\partial^2 \ln(p) / \partial \fone ^2 = -2$.
Hence, $\ln(p)$ is maximized by any values for $\fone$ and $\ftwo$ that satisfy:
\begin{equation}\label{eq:maximize-posterior}
     \fone + \ftwo = \g_a + \g_b,
\end{equation}
which also implies $\g_c = \frac{1}{2}(\g_a + \g_b)$.

We now see it is possible to have $\fone=\ftwo=\frac{1}{2}(\g_a + \g_b)$, i.e., both hypotheses are always the same, and always point to the ``midpoint'' between ``First'' and ``Fursten.''
In such a case, our network is equivalent to the single-embedding case, and it serves no purpose to produce two embeddings.

However, the analysis changes when we assume a high number of dimensions. The vectors become sparse, so we can no longer assume the existence of $\g_c$ in the vocabulary.
It is also known that the $\L_2$ distances become more and more similar so that there is little difference between ``near'' and ``far.'' \cite{beyer-1999}.
To simplify our analysis, in Appendix \ref{appendix-l2} we approximate all $\L_2$ distances in high dimensions as \emph{binary} distances, i.e., 0 if there is an exact match between the vectors, and some positive constant $D$ otherwise:
\begin{equation}\label{eq:approx_binary}
s_j = D \big\{ \delta(\fone - \g_j)-1 \big\} + D \big\{ (\delta(\ftwo - \g_j)-1 \big\} \ \ \forall j,
\end{equation}
where $\delta(.)$ is 1 if the input is 0, and 0 otherwise.
Applying Equation \eqref{eq:approx_binary} to \eqref{eq:joint-posterior}, we can identify 8 different cases for $\ln(p)$ after accounting for symmetry, shown in Appendix \ref{appendix-post-high}, to see that $\ln p$ is maximized when $\fone=\g_a$ and $\ftwo=\g_b$, or vice versa.
We therefore conclude that for a sufficiently high number of dimensions, the network will be trained so that $\fone$ becomes the embedding for ``First'' and $\ftwo$ becomes the embedding for ``Fursten'' (or vice versa).

\section{Proposed Method}
\label{sec:proposed-method}

\subsection{Basic Intuition}
\label{subsec:intuition}

The following two arguments are central to our approach:
\begin{enumerate}
\item CTC models do not give timely reports, but they can still report accurate timings. 
\item We use a vocabulary to train an embedding-matching A2W, but teaching the vocabulary is not our main goal.
\end{enumerate}
It is well-known that CTC \cite{graves-2006}, by design, operates in a ``segmentation-free'' manner.
The training criterion does not require segmentation information, nor is segmentation outputted by the model.
A CTC model's decision about a word can be made at any indeterminate point in time, as long as the correct words are produced in the right order.
This, however, does not necessarily mean that the model cannot \emph{know} the segmentation.
In an embedding-matching CTC model, we previously argued \cite{yen-2023} that the network acts as a word segmenter followed by an audio segment encoder.
If this interpretation is indeed true, then the CTC model \emph{must} internally know the precise location and duration of every word it hypothesizes, and we need only to find a way to expose that knowledge.

Consider an example vocabulary of 3 words, $\{I, like, honey\}$. 
We imagine expanding this to a vocabulary of \emph{timestamped words}, each with the form ``$\texttt{<word>\_<start time>\_<duration>}$''. 
If there are 100 possible start times and 10 possible durations, we would have a vocabulary of 3,000 timestamped words, i.e., $\{I\_1\_1, \cdots, I\_100\_10, like\_1\_1, \cdots, honey\_100\_10\}$.
Suppose we define the embedding vector for each timestamped word as the original word's embedding vector with the start time and duration appended as two additional dimensions.
If the text embedding for ``like'' is $\g_{like}$, the embedding for ``$like\_{\alpha}\_{\beta}$'' is $\g_{like\_{\alpha}\_{\beta}} = [\g_{like}^T \ \ \alpha \ \ \beta ]^T$.

It is then conceivable that we use a training method similar to Section \ref{sec:analysis1} with the 3-word vocabulary replaced by the expanded 3,000-word vocabulary.
Reference start times and durations for the training samples are prepared via forced alignment using a conventional DNN-HMM hybrid ASR \cite{mohamed-2012, mohri-2008, zhen-huang-2020}, and are attached to the words when training the CTC, e.g. ``\emph{I\_19\_2} \emph{like\_31\_5} \emph{honey\_52\_9}.''.
Once trained, the CTC model outputs sequences from the timestamped 3,000-word vocabulary, and we have achieved our goal.
Obviously, this approach is impractical. The training vocabulary is increased by a factor of 1,000, which is prohibitively large. 

This brings us to the second argument above.
Note that in our previous work \cite{yen-2023}, the training vocabulary was less than 1/10th of the testing vocabulary.
Our training goal is to make the conformer learn to segment the audio, and map (as an $f(\cdot)$ encoder in \cite{jeon-ane-2020} or \cite{he-2017}) each segment to an embedding.
As long as this goal is achieved, it doesn't matter what exact words we put into $G$ (the matrix of $\g_i$'s in Section \ref{subsec:single-embedding}) during training.
We could in fact put a \emph{random} subset of words into $G$ that changes for \emph{every} training sample. \footnote{Note that the model probably does learn some linguistic information that helps it identify word segments, but our argument still stands that as long as we keep the randomized $G$ reasonably large and fairly sampled, the exact words we include it in are not that important.}
In our hypothetical scenario, we need not use all 3,000 timestamped words in $G$ during training.
Rather, we can use a small random sample.
A static space of 1,000 possible timestamps does not even need to be defined;
one could use random start and duration values where needed, as we will describe in greater detail in the next section.

\subsection{Timestamped Single-Hypothesis System}
\label{subsec:ts-single-embedding}


\begin{figure}
     \centering
     \includegraphics[width=0.8\linewidth]{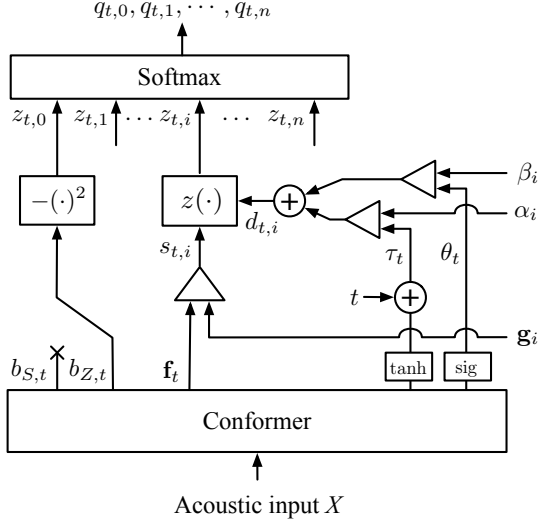}
     \caption{
     Training mode of the proposed single-hypothesis system; 
     in particular, the computation of the timestamped-word CTC loss $\L_Z$ at time $t$.
     A comparator ($\triangle$) represents the Euclidean distance between two inputs (for $s_{t,i}$, the distance is negated). 
     For clarity, we show only the computation of $z_{t,i}$ for the $i$'th word in the training vocabulary of size $n$.
     The same process is applied to all the other words.
     Note that $b_{S,t}$ is used for computing $\L_S$, so is unused here.
     Also note that $b_{Z,t}$, $z_{t,i}$, and $q_{t,i}$ are only used during training, and not during testing (shown in Fig. \ref{fig:ts-single-embedding-test}).
     }
     \label{fig:ts-single-embedding-train}
\end{figure}

\begin{figure}
     \centering
     \includegraphics[width=.8\linewidth]{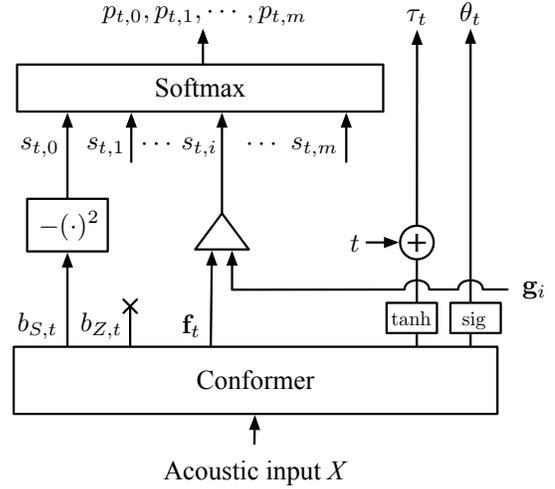}
     \caption{
     Testing mode of the proposed single-hypothesis system.
     We compute the $s_{t,i}$'s and $p_{t,i}$'s the same way as in our original (non-timestamped) embedding-matching system \cite{yen-2023} for every word in a testing vocabulary of size $m$, and also output a timestamp $(\tau_t, \theta_t)$ at every time $t$.
     The blank score $b_{Z,t}$ used to train $\L_Z$ is ignored during testing.
     }
     \label{fig:ts-single-embedding-test}
\end{figure}

We now formally describe the single-embedding version of our proposed system, illustrated in Fig. \ref{fig:ts-single-embedding-train} and \ref{fig:ts-single-embedding-test}.
At every time $t$, the underlying model proposes a word-based blank score $b_{S,t}$, a timestamped-word-based blank score $b_{Z,t}$, an acoustic embedding vector $\f_t$, and a timestamp vector $\u_t = [ \tau_t \ \ \theta_t ]^T$ where $\tau_t$ is the start time and $\theta_t$ is the duration for $\f_t$.

When producing $\tau_t$, the time $t$ of the current output frame is added first, so the neural network is actually trained to produce \emph{offsets} rather than absolute times.
A scaled hyperbolic tangent is used to limit the offset to $-1$ to $+1$ seconds. 
When producing $\theta_t$, a sigmoid is used to ensure that the value is positive, with a scale factor of 2 assuming a maximum word duration of 2 seconds.
In Appendix \ref{offset-duration-limits}, we show that both limits are conservatively high, and do not have any effect at testing time.
However, when the model was trained without the sigmoid and hyperbolic tangent, some degradation was observed in the recognition and timestamp accuracy, implying that the operations still help at training time.

The model is trained by using the sum of two CTC losses as the optimization criterion: a word CTC loss $\L_S$ and a timestamped-word CTC loss $\L_Z$:
\begin{equation} \label{eq:total_loss}
\L = \L_S + \L_Z.
\end{equation}
The word CTC loss $\L_S$ is exactly the same as that we used previously \cite{yen-2023}, defined in Equation \eqref{eq:ctc-loss}, using the scores in Equation \eqref{eq:score} where $-b^2_{S,t}$ is the blank score:
\begin{equation} \label{eq:ts-score}
s_{t,i} = 
\begin{cases}
\displaystyle -b_{S,t}^2 & (i=0) \\[1ex]
\displaystyle -\Vert\f_t - \g_i \Vert^2 & (i=1, \cdots, n)
\end{cases}
\end{equation}

The timestamped-word CTC loss $\L_Z$ includes a sequence of word timestamps $V$ corresponding to the word sequence $W$ of every training sample:
\begin{equation}
\L_Z = - \hspace{-10pt} \sum_{(X, W, V) \in \mathcal{D}} \hspace{-10pt} \ln \hat{P}(W, V|X).
\end{equation}
For computing $\hat{P}(W,V|X)$, we use the following pre-softmax score:
\begin{equation} \label{eq:ts-single-score}
    z_{t,i} = 
    \begin{cases}
    \displaystyle -b_{Z,t}^2 & (i=0) \\[1ex]
    \displaystyle z \left( s_{t,i}, d_{t,i} \right) & (i=1, \cdots, m)
    \end{cases}
\end{equation}
where
\begin{align} 
z(s, d) & \triangleq s - d + s d \label{eq:z_fun}\\
d_{t,i} & \triangleq \Vert \u_t - \v_i \Vert ^2 = (\tau_t - \alpha_i)^2 + (\theta_t - \beta_i)^2 . \label{eq:timestamp_error1}
\end{align}

Here, we do not use a fixed-vocabulary $G$.
A timestamped-vocabulary of size $m$ is prepared on-the-fly for every training sample.
Each $i$'th entry in the vocabulary is a tuple $(\g_i, \v_i)$ where $\v_i$ is a timestamp $[ \alpha_i \ \ \beta_i ]^T$for $\g_i$.
The $m$ training samples consist of reference samples and negative samples.
The reference samples are the reference words and their corresponding reference timestamps (obtained from forced-alignment using a DNN-HMM ASR).
The negative samples are generated by taking 1. All the reference words, with random perturbation added to their timestamps and 2. Randomly-chosen non-reference words with random timestamps, where the timestamps are perturbed versions of the timestamps in the reference transcription.
We found that such perturbed versions of the reference timestamps -- rather than purely random timestamps -- give better results.

The post-softmax score is 
\begin{equation}
    q_{t,i} = \frac{\exp(z_{t,i})}{\sum_j \exp(z_{t,j})}.
\end{equation}
By inspection, we can see that the same argument can be made regarding $q_{t,i}$ as we did for $p_{t,i}$ in Sec. \ref{subsec:single-embedding}.
Since $d_{t,i} \ge 0$, it always makes $z_{t,i}$ lower, and leaves $z_{t,i}$ unchanged when $d_{t,i}=0 \leftrightarrow \u_t=\v_i$.
When $\f_t=\g_i$, $\tau_t = \alpha_i$, and $\theta_t = \beta_i$, we get the maximum value for $z_{t,i}$, and for a sufficiently large number of dimensions in the embedding vectors, $z_{t,i}$ will be much higher than the other $z_{t,j}$'s, allowing $q_{t,i}$ to be close to the maximum value of 1. 

This also implies that $\L_Z$ may be all we need in Equation \eqref{eq:total_loss}.
The reason why we still include $\L_S$ is the blank score.
$-b_{Z,t}^2$ from $\L_Z$ is trained to compete with the non-blank scores in Equation \eqref{eq:ts-single-score}. However, at testing time, we use the non-blank scores in Equation \eqref{eq:score}, so must use a compatible blank score, which is $-b_{S,t}^2$ trained using $\L_S$.
Also note that the non-blank scores in Equation \eqref{eq:ts-single-score} are generally lower than the non-blank scores in Equation \eqref{eq:score} because the former are weighed down by the timestamp estimation errors.
We would get many insertion errors if we used $-b_{Z,t}^2$ for testing, because it must compete with non-blank scores that are overall higher than what it had seen at training time.

Testing mode is shown in Fig. \ref{fig:ts-single-embedding-test}.
When testing, we use $s_{t,i}$ and $p_{t,i}$ in Equation \eqref{eq:score} and \eqref{eq:softmax} as in the previous system, and attach a timestamp vector $\u_{t,i} = [\tau_t \ \ \theta_t]^T$ to every $s_{t,i}$ $(i>0)$.
Whenever the prefix beam decoder chooses a word $w_i$ at time $t$ with score $s_{t,i}$, it takes $\u_{t,i}$ as the word's timestamp.

\subsection{Timestamped Multiple-Hypothesis System}
\label{subsec:ts-multiple-embedding}

\DeclareRobustCommand{\trapezoid}{
\begin{tikzpicture}
    \node[trapezium,
     draw,
     minimum width = 1pt,
     minimum height= 1pt,
     trapezium stretches = true,
    ] (t) at (0,0) {};
\end{tikzpicture}
}

\begin{figure}[t]
     \centering
     \includegraphics[width=.95\linewidth]{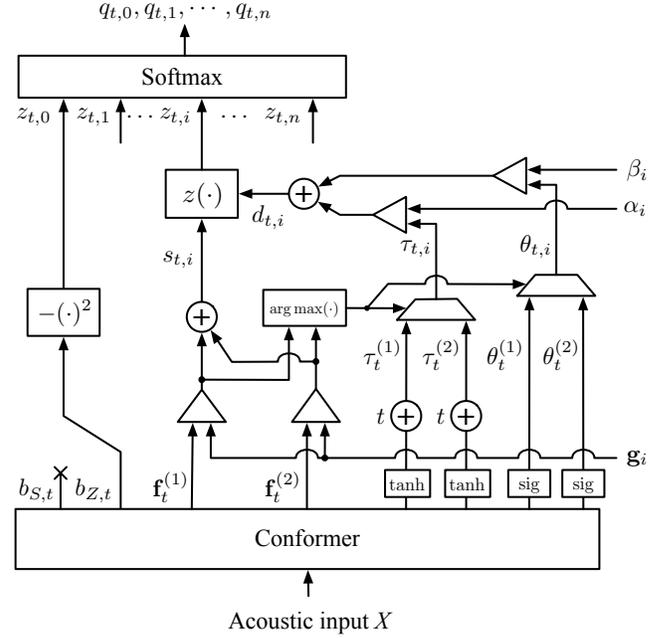}
     \caption{
          Training mode of the proposed system when $L=2$; in particular, computation of $\L_Z$ at time $t$. 
          A comparator ($\triangle$) represents the $\L_2$ distance (negated for $s_{t,i}$) between two inputs.
          A multiplexer (\hspace{-.2em}\trapezoid\hspace{-.2em}) chooses one of two timestamps based on the hypothesized embedding that best matches the $i$'th word.
          For clarity, we show only the computation of $z_{t,0}$ and $z_{t,i}$ in this figure.
          The other $z_{t,j}$'s are computed the same way, using different $\alpha_j$, $\beta_j$ and $\g_j$.
          Note that $b_{S,t}$ is used for computing $\L_S$, so is unused here.
     }
     \label{fig:ts-multiple-embedding-train}
\end{figure}

\begin{figure}
     \centering
     \includegraphics[width=.95\linewidth]{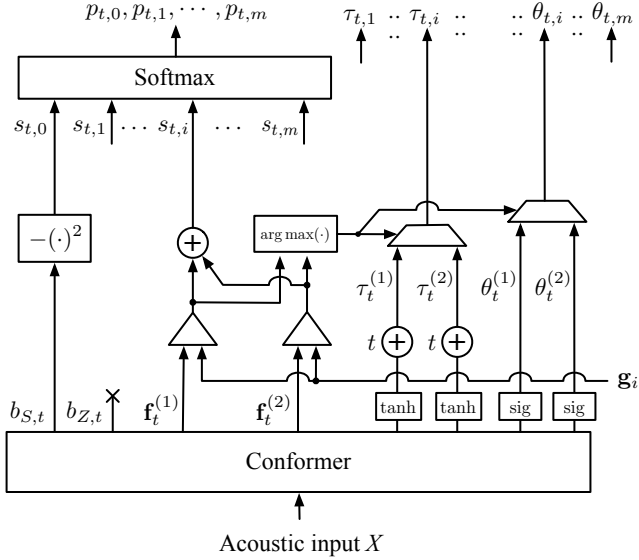}
     \caption{
     Testing mode of the proposed multiple-hypothesis ($L=2$) system.
     At time $t$, for every word $w_i$ in the testing vocabulary, the network produces a posterior score $p_{t,i}$, a start time $\tau_{t,i}$, and a duration $\theta_{t,i}$.
     For clarity, in the figure we only show the score computation for the $i$'th word.
     The same process is applied to all the other words.
     The blank score $b_{Z,t}$ used to train $\L_Z$ is ignored during testing.
     }
     \label{fig:ts-multiple-embedding-test}
\end{figure}

The multiple-embedding version of our system is shown in Fig. \ref{fig:ts-multiple-embedding-train} and Fig. \ref{fig:ts-multiple-embedding-test}.
For each $k$'th embedding vector $\f_t^{(k)}$, the conformer outputs a timestamp vector $\u_t^{(k)} = [ \tau_t^{(k)} \ \ \theta_t^{(k)} ]$.

For each label $i$, the pre-softmax score for $\L_Z$ is based on the multiple-embedding score $s_{t,i}$ in Equation \eqref{eq:logical-or} and the hypothesized timestamp corresponding to the best-matching hypothesized embedding:
\begin{equation}\label{eq:ts-multi-score}
    z_{t,i} = 
    \begin{cases}
     \displaystyle -b_{Z,t}^2 & (i=0) \\[1ex]
    \displaystyle z(s_{t,i}, d_{t,i}) & (i=1, \cdots, n),
    \end{cases}
\end{equation}
where $z(\cdot)$ is defined in Equation \eqref{eq:z_fun} and
\begin{align}
    s_{t,i} & = \textstyle { \sum_{k=1}^L } s_{t,i}^{(k)} \ \ (i=1, \cdots, n) \label{eq:ts-multi-score-s}\\
    s_{t,i}^{(k)} & = - \Vert\f_t^{(k)} - \g_i \Vert^2 \ \ (i=1, \cdots, n) \label{eq:ts-multi-score-sk}\\
    d_{t,i} & = \Vert \u_{t,i} - \v_i \Vert ^2 = (\tau_{t,i} - \alpha_i)^2 + (\theta_{t,i} - \beta_i)^2 \label{eq:ts-multi-score-d}\\
    \u_{t,i} & = \u_t^{(l)} \\
    l & = \arg \max_k s_{t,i}^{(k)} \ \ (i=1, \cdots, n) . \label{eq:ts-multi-score-l}
\end{align}

The reasoning behind Equation \eqref{eq:ts-multi-score-l} is that since each $\fk$ corresponds to one segment in the speech, for every word $w_i$ it will be the $\fk$ that best matches the word that will also have the most accurate timestamp.
We only care about cases when there exists a strong match with $w_i$.
If all $\fk$'s have a weak match with $w_i$, the word wouldn't be chosen by the decoder so it doesn't matter what timestamp we assign to it.
If multiple $\fk$'s have a strong match with the same $w_i$, those $\fk$'s are also likely to have very similar timestamps (since they are representing the same audio segment), so we could choose any one of them.

\subsection{Analysis of Timestamped Multiple-Hypothesis System}
\label{subsec:analysis-ts-multiple-embedding}

\begin{figure}
     \centering
     \includegraphics[width=1.05\linewidth]{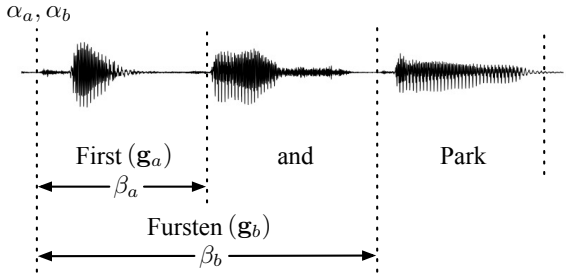}
     \caption{
     Extension of the example in Fig. \ref{fig:fursten-wav1} with timestamps $\v_a=(\alpha_a, \beta_a)$ and $\v_b=(\alpha_b, \beta_b)$ included in the reference transcriptions where $\alpha_i$ and $\beta_i$ are the start time and duration, respectively, of word $w_i$.
     This time, for one audio input, the multiple-hypothesis embedding-matching system must learn to produce two different \emph{timestamped} embeddings $(\g_a, \v_a)$ and $(\g_b, \v_b)$.
     }
     \label{fig:fursten-wav2}
\end{figure}

We extend the example in Sec. \ref{subsec:multiple-embedding} to assume one input training audio with two different reference \emph{timestamped} word transcriptions.
We want to show that the timestamped multiple-hypothesis system with $L=2$ will be able to separately represent \emph{both} the word embeddings and the timestamps for ``First'' and ``Fursten.''

Word $w_a$ (``First'') is described by $(\g_a, \v_a)$ where $\v_a$ stores the reference timestamp $[ \alpha_a, \beta_b ]^T$.
Word $w_b$ (``Fursten'') is described by $(\g_b, \v_b)$. The CTC's underlying model produces two embeddings $\f_t^{(1)}$ and $\f_t^{(2)}$, and two corresponding timestamps $\u_t^{(1)} = [\tau_t^{(1)}, \theta_t^{(1)}]$ and $\u_t^{(2)}= [\tau_t^{(2)}, \theta_t^{(2)}]$.

Since our optimization criterion in \eqref{eq:total_loss} is the sum of the log of two CTC losses, we can consider the product of the post-softmax scores $p_a \cdot p_b \cdot q_a \cdot q_b$ at some point in time $t$ that CTC training would attempt to maximize.
We will show that the following condition is a sufficient condition for maximizing $q$:
\begin{equation}\label{eq:maximize-condition}
\f_t^{(1)} = \g_a, \ \
\u_t^{(1)} = \v_a,\ \
\f_t^{(2)} = \g_b,\ \
\textrm{and} \ \ 
\u_t^{(2)} = \v_b
\end{equation}
where 1 and 2 may be swapped.

We already know that \eqref{eq:maximize-condition} maximizes $p_a \cdot p_b$, so we only consider $q = q_a \cdot q_b$. 
As before, we only need to consider 1 point in time, so we drop $t$ to simplify the notation. 
Note that $\fone=\g_a$ implies $\u_{a}=\u^{(1)}$, i.e., for word $w_a$, the first embedding is the obvious ``winner'', so $w_a$ is also assigned the first timestamp.
Likewise, $\ftwo=\g_b$ implies $\u_{b}=\u^{(2)}$.

First, one can see that
\begin{equation}
     \frac{\partial z_j}{\partial \u_i} = 
     \begin{cases}
          2 \left( \u_i - \v_i \right) (s_i - 1), & \textrm{for } i=j \\
          0,                                      & \textrm{for } i \ne j .
     \end{cases}
\end{equation}
We wish to maximize
\begin{equation}
\ln(q)=\ln(q_a) + \ln(q_b)=z_a + z_b - 2\ln \sum_j \exp(z_j).
\end{equation}
We can see that
\begin{align}
\frac{\partial \ln q}{\partial \u_a}
& = \frac{\partial z_a}{\partial \u_a} \left(1 - 2 q_a \right) \\
\frac{\partial^2 \ln q}{\partial \u^2_a}
& = 2(s_a - 1) \left(1 - 2 q_a \right)
-2 \left( \frac{\partial z_a}{\partial \u_a} \right)^2 q_a(1-q_a) .
\end{align}
Hence, $q_a < 1/2$ is a sufficient condition for $\ln q$ to be concave with respect to $\u_a$, in which case $\ln q$ will be maximized with respect to $\u_a$ when $\u_a = \v_a$.
By symmetry, if $q_b < 1/2$, $\ln q$ will be maximized with respect to $\u_b$ when $\u_b = \v_b$.
In Appendix \ref{appendix-labor2}, we apply the same assumption as in Sec. \ref{subsec:multiple-embedding} for high embedding dimensions to show that $q_a < 1/2$ and $q_b < 1/2$ indeed holds, and we also show that $\fone=\g_a$ and $\ftwo=\g_b$ maximize $q$. Hence, Condition \eqref{eq:maximize-condition} is a sufficient condition for maximizing $q$.

\section{Experiments}

We trained ``previous'' embedding-matching models using word-pronunciation embeddings in the same manner as we previously proposed in \cite{yen-2023}.
The training and development data consisted of utterances spoken by diverse anonymized speakers to a digital assistant.
The training data had a total 2M utterances (3,030 hours), and the development data had 400k utterances (534 hours).
We then trained the proposed timestamped models using the same audio and embeddings as the previous models, but also with ground truth timestamps provided by a DNN-HMM ASR.
We also varied the number of internal embeddings (the value of $L$ in Equations \eqref{eq:logical-or} and \eqref{eq:ts-multi-score-s}) from 1 to 3.

All models were conformer \cite{gulati-2020} networks, with 272 encoder dimensions, 8 attention heads, 18 layers, and a kernel size of 31.
A final linear transformation was applied to the conformer's output to obtain the desired number of embeddings and the timestamps (when applicable).
For example, for the proposed system with 3 internal embeddings, the final output of the model had $40 \times 3 + 2 + 2 \times 3=127$ dimensions (3 embeddings, 2 blank scores, 3 start times, and 3 durations).
The final models had the following total number of parameters:
\begin{table}[h]
  \caption{Total number of trainable model parameters}
  \label{tab:models}
  \centering
  \begin{tabular}{lcc}
    \toprule
    \multirow{2}{*}{Model} & Internal & \multirow{2}{*}{Parameters} \\
                           & Embeddings ($L$)  & \\
    \midrule
    \multirow{3}{*}{Previous}   & 1 & 32,403,673 \\
       & 2 & 32,414,593 \\
       & 3 & 32,425,513 \\
    \hline
    \multirow{3}{*}{Proposed}   & 1 & 32,404,492 \\
       & 2 & 32,415,958 \\
       & 3 & 32,427,424 \\
    \bottomrule
  \end{tabular}
\end{table}

The training schedule follows the 2-phase specification in \cite{gulati-2020} and \cite{vaswani-2017}, but with an additional third phase where the learning rate decays exponentially (rather than following the inverse square root), which we found to improve accuracy.
The schedule is specified by 4 variables: the number of steps $p_1$ in the first phase (warmup phase), the number of steps $p_2$ in the second phase (inverse square root), the exponential decay rate $\gamma$ in the third phase, and the peak learning rate $R$.
The learning rate at training step $s$ is
\begin{equation} \label{eq:learning_rate}
\begin{dcases}
\displaystyle \frac{R}{p_1} s & 1 \le s \le p_1 \\
\displaystyle \frac{R \sqrt{p_1}}{\sqrt{s}} & p_1 \le s \le p_1 + p_2 \\
\displaystyle R \sqrt{\frac{p_1}{p_1+p_2}} \cdot 2 ^ { ( p_1 + p_2 )/ \gamma} \cdot 2 ^ {- s/\gamma} & p_1 + p_2 \le s
\end{dcases}
\end{equation}

We used $R=0.01 / \sqrt{d}$ where $d$ is the encoder dimensions of the conformer.
We also set $p_1=30$k, $p_2=270$k, and $\gamma=30$k with a batch size of 256.
A convolutional subsampling was applied to the acoustic input before the conformer, resulting in a frame rate of 25 frames/second (or 40ms per frame).
For every training job, the model produced at every epoch was evaluated on a small held-out tuning dataset of 4k utterances, and the model that had the lowest WER was chosen as the final model.

All embedding-matching CTC models were word-pronunciation-based models as described in \cite{yen-2023}, using a pronunciation-based text encoder as described in \cite{jeon-ane-2020} that outputs text embeddings with 40 dimensions.
For the previous models, the training vocabulary size was 39,478.
For the timestamped models, the randomized timestamped vocabulary size was fixed to a similar size, 40,000.

\subsection{Segmentation Accuracy on TIMIT Data}
\label{subsec:segmentation}

\newcolumntype{C}[1]{>{\centering\arraybackslash}p{#1}}
\begin{table*}
\centering
\caption{Labels outputted by the proposed model at every 40ms frame for $\tt{dr8/mres0/si1847}$, starting at frame 0 (the top left cell), and ending at frame 58. Each row in the table shows 10 frames (the first row shows frames 0 $\sim$ 9, the second row shows 10 $\sim$ 19, and so on).}
\label{tab:si1847-labels}
\setlength{\tabcolsep}{3pt}
\begin{tabular}
{c  |   C{1cm} C{1cm} C{1cm} C{1cm} C{1cm} C{1cm} C{1cm} C{1cm} C{1cm} C{1cm} }
\hline
Frame No. & 0  & 1  & 2  & 3  & 4 & 5  & 6  & 7 & 8  & 9 \\
\hline
0  & \_ & \_ & \_ & \_ & \_ & she & \_ & \_ & \_ & \_ \\
10 & \_ & \_ & radiated & radiated & \_ & \_ & \_ & \_ & \_ & \_ \\
20 & \_ & \_ & \_ & \_ & \_ & \_ & warmth & \_ & \_ & \_ \\
30 & \_ & \_ & \_ & \_ & \_ & \_ & \_ & and & \_ & \_ \\
40 & \_ & good & \_ & \_ & \_ & \_ & \_ & \_ & \_ & \_ \\
50 & fellowship & \_ & \_ & \_ & \_ & \_ & \_ & \_ & \_ \\
\hline
\end{tabular}
\end{table*}

\begin{table*}
\centering
\caption{Word start times (in ms) produced by proposed model for the label sequence in Table. \ref{tab:si1847-labels}. For the repeated label at frames 12 and 13, the word start times are near-identical. The corresponding reference start times (included in the TIMIT data) are 141, 362, 992, 1512, 1649, and 1864.}
\label{tab:si1847-start}
\setlength{\tabcolsep}{3pt}
\begin{tabular}
{c  |   C{1cm} C{1cm} C{1cm} C{1cm} C{1cm} C{1cm} C{1cm} C{1cm} C{1cm} C{1cm} }
\hline
Frame No. & 0  & 1  & 2  & 3  & 4 & 5  & 6  & 7 & 8  & 9 \\
\hline
0  & \_ & \_ & \_ & \_ & \_ & 84 & \_ & \_ & \_ & \_ \\
10 & \_ & \_ & 388 & 389 & \_ & \_ & \_ & \_ & \_ & \_ \\
20 & \_ & \_ & \_ & \_ & \_ & \_ & 1024 & \_ & \_ & \_ \\
30 & \_ & \_ & \_ & \_ & \_ & \_ & \_ & 1469 & \_ & \_ \\
40 & \_ & 1573 & \_ & \_ & \_ & \_ & \_ & \_ & \_ & \_ \\
50 & 1875 & \_ & \_ & \_ & \_ & \_ & \_ & \_ & \_ \\
\hline
\end{tabular}
\end{table*}

\begin{table*}
\centering
\caption{Durations (in ms) produced by proposed model for the label sequence in Table. \ref{tab:si1847-labels}. For the repeated label at frames 12 and 13, the durations are near-identical. The corresponding reference durations (included in the TIMIT data) are 221, 630, 520, 137, 215, and 587.}
\label{tab:si1847-end}
\setlength{\tabcolsep}{3pt}
\begin{tabular}
{c  |   C{1cm} C{1cm} C{1cm} C{1cm} C{1cm} C{1cm} C{1cm} C{1cm} C{1cm} C{1cm} }
\hline
Frame No. & 0  & 1  & 2  & 3  & 4 & 5  & 6  & 7 & 8  & 9 \\
\hline
0  & \_ & \_ & \_ & \_ & \_ & 270 & \_ & \_ & \_ & \_ \\
10 & \_ & \_ & 597 & 599 & \_ & \_ & \_ & \_ & \_ & \_ \\
20 & \_ & \_ & \_ & \_ & \_ & \_ & 423 & \_ & \_ & \_ \\
30 & \_ & \_ & \_ & \_ & \_ & \_ & \_ & 73 & \_ & \_ \\
40 & \_ & 256 & \_ & \_ & \_ & \_ & \_ & \_ & \_ & \_ \\
50 & 600 & \_ & \_ & \_ & \_ & \_ & \_ & \_ & \_ \\
\hline
\end{tabular}
\end{table*}

\begin{table}[t]
  \caption{Word segmentation accuracy on TIMIT data (lower values are better). 
  The mean $\mu$ in milliseconds of the absolute difference between the hypothesized and reference start times ($\alpha$) and durations ($\beta$) are shown for a hybrid DNN-HMM ASR and the proposed timestamped system.
  }
  \label{tab:timit}
  \centering
  \begin{tabular}{lcccc}
    \toprule
    \multirow{2}{*}{System} & Internal & \multirow{2}{*}{$\mu_\alpha$ (ms)} & \multirow{2}{*}{$\mu_\beta$} \\
           & Embeddings & & \\
    \midrule
    Baseline DNN-HMM & (Not applicable) & 18.9  & 30.4  \\
    \hline
    \multirow{3}{*}{\shortstack[l]{Proposed \\(trained on 3k hours)}}  & 1 & 23.6  & 38.7 \\
     & 2 & 22.6  & 37.4 \\
     & 3 & 21.8  & 37.1 \\
    \bottomrule
  \end{tabular}
\end{table}

\begin{table}[t]
  \caption{Word segmentation accuracy on Noisy TIMIT data (lower values are better). 
  The mean $\mu$ in milliseconds of the absolute difference between the hypothesized and reference start times ($\alpha$) and durations ($\beta$) are shown for a hybrid DNN-HMM ASR and the proposed timestamped system using 3 internal embeddings.
  }
  \label{tab:noisy_timit}
  \centering
  \begin{tabular}{ccccc}
    \toprule
    System & Noise Type & SNR & $\mu_\alpha$ (ms) & $\mu_\beta$ (ms)\\
    \midrule
    \multirow{12}{*}{\shortstack[c]{Baseline \\DNN-HMM}} & \multirow{4}{*}{Babble} & 20 & 20.5  & 32.2  \\
                                       &                         & 15 & 21.6  & 34.6 \\
                                       &                         & 10 & 24.0  & 40.9 \\
                                       &                         & 5  & 31.4  & 57.3 \\
                                       \cline{2-5}
                                       & \multirow{4}{*}{Pink} & 20 & 18.5  & 31.4  \\
                                       &                         & 15 & 19.8  & 35.7 \\
                                       &                         & 10 & 23.6  & 46.3 \\
                                       &                         & 5  & 33.7  & 69.1 \\
                                       \cline{2-5}
                                       & \multirow{4}{*}{White} & 20 & 19.4  & 34.6 \\
                                       &                         & 15 & 21.8  & 42.4 \\
                                       &                         & 10 & 27.2  & 56.6 \\
                                       &                         & 5  & 39.1  & 81.5 \\
    \midrule
    \multirow{12}{*}{\shortstack[c]{Proposed \\ ($L=3$)}} & \multirow{4}{*}{Babble} & 20 & 23.3  & 37.8 \\
                                       &                         & 15 & 24.8  & 40.1 \\
                                       &                         & 10 & 28.3  & 47.9 \\
                                       &                         & 5  & 41.4  & 70.4 \\
                                       \cline{2-5}
                                       & \multirow{4}{*}{Pink} & 20 & 21.5  & 36.4 \\
                                       &                         & 15 & 22.9  & 39.7 \\
                                       &                         & 10 & 26.9  & 49.5 \\
                                       &                         & 5  & 39.2  & 74.8 \\
                                       \cline{2-5}
                                       & \multirow{4}{*}{White} & 20 & 22.2  & 38.0 \\
                                       &                         & 15 & 24.4  & 43.6 \\
                                       &                         & 10 & 30.3  & 57.0 \\
                                       &                         & 5  & 48.6  & 89.1 \\
    \bottomrule
  \end{tabular}
\end{table}

The TIMIT \cite{garofolo-1992} dataset includes human-edited reference word timestamps \cite{zue-1996}.
We force-aligned the TIMIT audio to the reference words using the proposed timestamped model, as well as a hybrid DNN-HMM ASR (the same model that provides the ground truth when training the proposed model) to obtain hypothesized start times and durations.
Force alignment using the embedding-matching models is done by constraining the prefix beam decoder (e.g. \cite{hannun-2014}) to only follow the reference words.
We compute the word start error as the absolute difference between the hypothesized start and the reference start for every word, then compute the mean of this error over all the words.
In another study \cite{yang-2022}, the same metric is called ``word absolute start time delta (WASTD).''
The same mean absolute error can be computed for word duration.

We will now describe how the decoder extracts word timestamps with the proposed CTC model.
The \emph{first} occurrence of each word in the 1-best alignment is used to determine the timestamp.
For example, assume the optimal label sequence found by the beam decoder has three words $w_{41}$=``how's'', $w_{25}$=``the'', and $w_{73}=$``weather'' as follows (``\_'' indicates blank):
\begin{center}
\small{
\setlength{\tabcolsep}{3pt}
\begin{tabular}
{l  |   c    c    c       c       c    c     c    c         c    c   c}
\hline
Time  & 0  & 1  & 2     & 3     & 4  & 5  & 6   & 7       & 8  & 9  & 10 \\
\hline
Word  & \_ & \_ & how's & how's & \_ & \_ & the & weather & \_ & \_ & \_ \\
\hline
\end{tabular}
}
\end{center}
The timestamps are 
$\u_{2,41}$, $\u_{6,25}$, $\u_{7,73}$, 
or $(\tau_{2,41}, \theta_{2,41})$, 
$(\tau_{6,25}, \theta_{6,25})$, 
$(\tau_{7,73}, \theta_{7,73})$.
Hence, the timestamp $\u_{3,41}$ associated with the second \emph{how's} is ignored.
Note, however, that $\u_{3,41}$ is most likely very similar to $\u_{2,41}$, because the timestamp is part of the word identity during CTC model training (the training wants \emph{both} $\u_t=\v_i$ \emph{and} $\f_t=\g_i$ in Section \ref{subsec:ts-single-embedding}).

Label repetitions actually occur rarely, but we show a real example from the TIMIT data that demonstrates the aforementioned effect.
For the $\tt{dr8/mres0/si1847}$ utterance, our decoder using the proposed timestamped CTC model produced the sequence of labels shown in Table \ref{tab:si1847-labels}, where the label for ``radiated'' was repeated in frames 12 and 13.

The word start and durations (in milliseconds) produced by the neural network for the corresponding labels are shown in Tables \ref{tab:si1847-start} and \ref{tab:si1847-end}, respectively.
For the repeated label, we can see that the start times and end times are near-identical for both frames 12 and 13.

A total 6,300 utterances containing a total 54,387 words were tested.
Table \ref{tab:timit} shows the results for the baseline hybrid DNN-HMM ASR, and the proposed timestamped embedding-matching CTC ASR using a variable number of internal embeddings.
The mean word start accuracy $\mu_\alpha$ (18.9ms for the baseline, 21.8ms for the proposed with 3 embeddings) is similar to the same measurement made for a GMM-HMM system in \cite{yang-2022}, which was 21.3ms.
Compared to the baseline hybrid DNN-HMM ASR, the proposed system has less than 3ms difference in average word start time accuracy, and less than 7ms difference in average word duration accuracy.
Note that the accuracy of the proposed model is bounded by the accuracy of the DNN-HMM, since the latter provides the ground truth timestamps when the former is trained.
The segmentation accuracy of the proposed system also improves with increasing number of internal embeddings.
This is probably for similar reasons as the improved recognition accuracy observed in \cite{yen-2023}, i.e., the ability of the embedding-matching model to generate more diverse hypotheses over more diverse word segments allow it to better represent the reference word sequences during forced alignment.

In Table \ref{tab:noisy_timit}, we ran the same experiment on Noisy TIMIT \cite{abdulaziz-2017} for three different types of noise with varying SNR.
For both the baseline DNN-HMM and the proposed systems, the accuracy degraded as the SNR decreased.
The accuracy gap between the baseline and the proposed systems also seemed to get wider for lower SNR, but in most cases, the gap was not larger than 10ms.

\subsection{Impact on Recognition Accuracy}
\label{subsec:recognition}

\setlength{\tabcolsep}{4pt}
\begin{table}[t]
  \caption{Recognition error rate (\% word error rate and named entity error rate) on digital assistant data with dynamic out-of-vocabulary words (contact names) for previous method \cite{yen-2023} and the proposed method.
  }
  \label{tab:assistant}
  \centering
  \begin{tabular}{lccc}
    \toprule
    \multirow{2}{*}{System}      & Internal & \multirow{2}{*}{WER}  & \multirow{2}{*}{NEER} \\
                & Embeddings ($L$) & & \\
    \midrule
    \multirow{3}{*}{Previous} & 1 & 9.5 & 9.7 \\
                & 2 & 9.2 & 9.3 \\
                & 3 & 8.9 & 9.0 \\
    \hline
    \multirow{3}{*}{Proposed} & 1 & 9.9 & 10.0 \\
                & 2 & 9.8 & 10.0 \\
                & 3 & 9.3 & 9.5  \\
    \bottomrule
  \end{tabular}
\end{table}

To test for any negative impact on recognition accuracy by the addition of $\L_Z$ in Equation \eqref{eq:total_loss}, we tested the system on a dataset of 35,907 utterances spoken by diverse anonymized speakers to a digital assistant.
For every utterance, a user-dependent list of contact names was added to the base vocabulary.
A language model (LM) that supports a \emph{\$CONTACT} variable was used with the decoder, where \emph{\$CONTACT} appears in contexts such as ``\emph{call \$CONTACT office}'' or \emph{``text \$CONTACT I'll be five minutes late.''}.
Table \ref{tab:assistant} shows the ASR accuracy.
In addition to the word error rate (WER), we also report a named entity error rate (NEER), which is the word error rate computed on only the named entity portions of the utterances.
When comparing the proposed system with the previous system in \cite{yen-2023} -- both using 3 internal embeddings -- we can notice a 4.5\% relative degradation of the WER and 5.5\% degradation of the NEER.
This is understandable given that both systems are using the exact same training data and nearly identical model size, but the proposed system has an additional term $\L_Z$ in the loss function that increases the complexity of the learning criterion and requires more information (the timestamps) to be modeled compared to the previous system.

For both the previous and proposed systems, there were 3 tunable parameters for each recognizer:
a \emph{named entity weight} that is applied to the acoustic score of any named entity hypothesized by the acoustic model, a \emph{language model weight} applied to the language model scores, and a constant \emph{blank score divisor} \cite{sak-2015} applied to the acoustic model's blank score.
The prefix beam decoder had 3 beam widths: an \emph{input beam} that limits the number of words emitted by the acoustic model at every frame, a \emph{label beam} that limits the total number of label sequences (including blanks) hypothesized by the decoder for each unique word sequence, and a \emph{word beam} that limits the total number of unique word sequences hypothesized by the decoder.
The label beam is always set to 1.
On a separate tuning dataset of 12k utterances, we set the input beam to 40 and the word beam to 100 and did a coarse parameter sweep to find the named entity weight, language model weight, and blank score divisor that minimized the word error rate.
These tuning parameters were then applied to the 36k evaluation data, using an input beam of 200 and a word beam of 500, to produce the results in Table \ref{tab:assistant}.

\begin{figure}[h]
\centering
     \subfloat[]{
          \vspace{0.2in}
          \includegraphics[width=\linewidth]{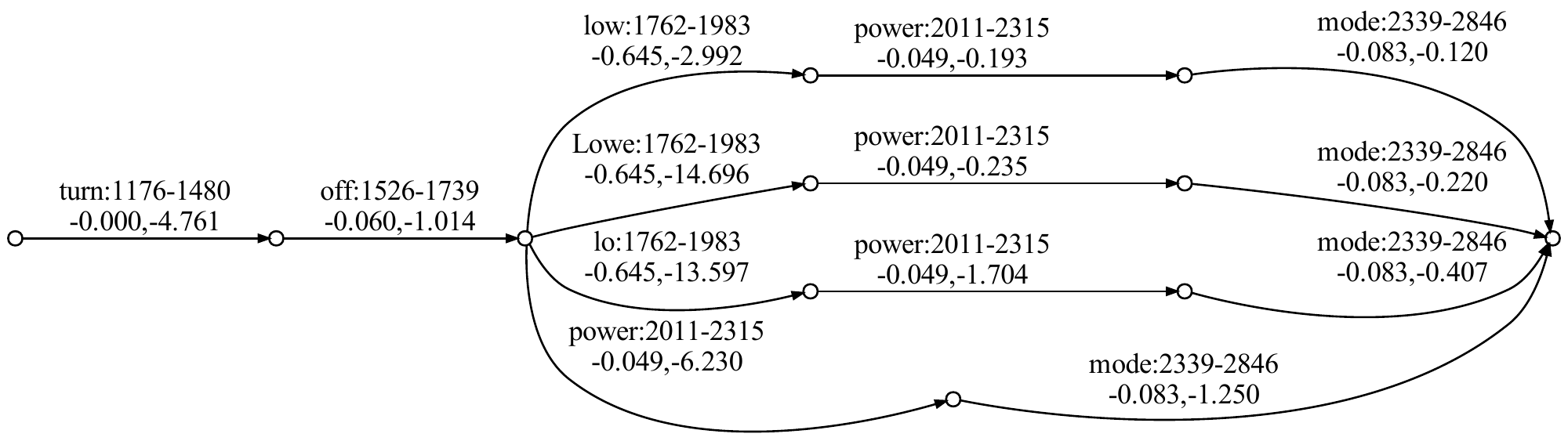}
          \label{fig:low-power-lattice}
     }
     \newline
     \subfloat[]{
     \vspace{0.2in}
          \includegraphics[width=\linewidth]{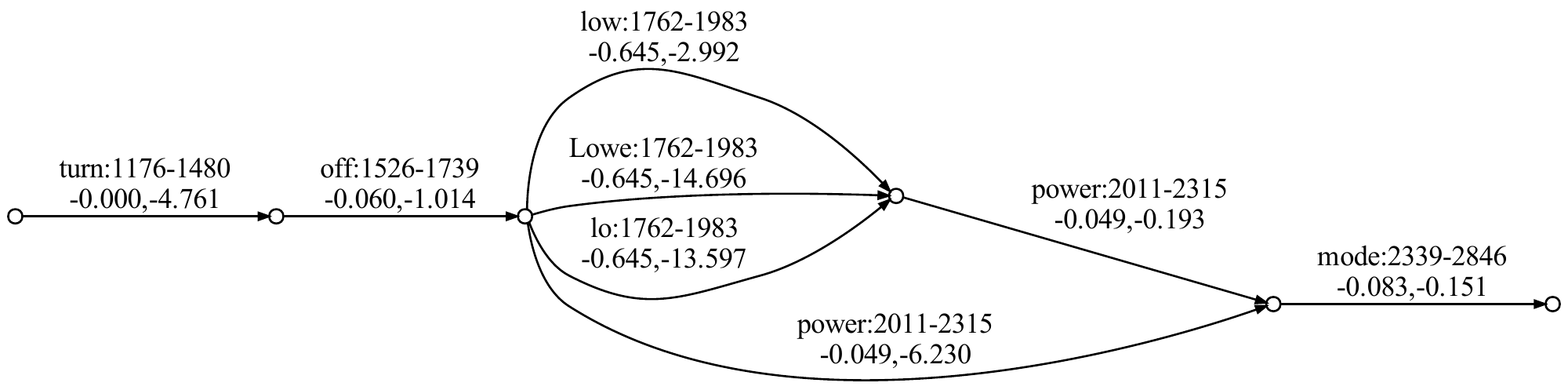}
          \label{fig:low-power-hwcn}
     }
     \caption{
     (a) Word hypothesis lattice outputted by the decoder for the utterance, ``\emph{turn off low power mode,}'' and (b) the corresponding heterogeneous word confusion network \cite{jeon-conf-2020}.
     Each arc is labeled as ``\texttt{<word>:<start time in ms>-<end time in ms> <acoustic score>, <LM score>}''.
     }
     \label{fig:low-power}
\end{figure}

In Appendix \ref{appendix-insertion}, we also discuss some (infrequent) cases where the timestamps allow the decoder to avoid insertion errors when the acoustic model hypothesizes words whose pronunciations partially overlap with each other.
The timestamps also enable the decoder to generate heterogeneous word confusion networks (HWCNs) \cite{jeon-conf-2020} for downstream processing.
The prefix tree built during decoding directly becomes the ASR's word hypothesis lattice, which can then be converted to an HWCN.
We show examples in Figure \ref{fig:low-power}.



\section{Conclusion and Future Work}
\label{sec:conclusion}

In this paper, a novel timestamped embedding-matching acoustic-to-word (A2W) model has been proposed, where a word-level embedding-matching CTC model is trained to directly estimate a start time and duration of every hypothesized word, allowing simpler, faster, and more efficient ASR.
Experiments show that the word segmentation accuracy can approach that of a hybrid DNN-HMM ASR, with small impact on word error rate.
We also contributed more rigorous analysis on multiple-hypothesis embedding-matching A2W to provide further insight into how the model can learn to produce multiple embeddings, and extended the analysis to show how the model can also produce word timestamps.

\section*{Acknowledgements}
\label{sec:ack}
Thanks to Tatiana Likhomanenko for helpful comments.

\appendices
\section{Maximization of the Softmax When Using the Inner Product} \label{appendix-a}

In the inner product case \cite{settle-2019,collobert-2020}, we have
\begin{equation}\label{eq:inner}
s_{i,t} = \f_t ^T \g_i .
\end{equation}

For embedding-matching to work, we need $s_{i,t}$ to be maximized if and only if $\f_t=\g_i$.
Such a property already always holds for the negative $\L_2$ distance in Equation \eqref{eq:score}, but for the inner product, we also require the normalization
\begin{equation}\label{eq:constraint}
\f_t^T \f_t = \g_i ^T \g_i = c \ \ \forall t, i ,
\end{equation}
for some constant $c>0$, so that by the Cauchy-Schwarz Inequality we would have
\begin{equation}\label{eq:constraint2}
(\f_t^T \g_i)^2 \le (\f_t^T \f_t)(\g_i^T \g_i) = c^2,
\end{equation}
where equality holds if and only if $\f_t=\g_i$.

If we were to take the approach of Sec. \ref{subsec:single-embedding}, we have
\begin{equation}\label{eq:log_softmax_them}
\ln p_{t,i} = -\ln \left[ 1 + \sum_{j \ne i}^n \exp \left ( \f_t^T \g_j - \f_t^T \g_i \right) \right] .
\end{equation}

We can apply the fact that any two isotropic random vectors on a $d$-dimensional sphere tend to be orthogonal as $d \to \infty$ \cite{vershynin-2018}. Setting $\f_t ^T \g_j=0$ and $\f_t ^T \g_i = c$ (since $\f_t=\g_i$), we obtain:
\begin{equation}\label{eq:log_softmax_them_bound}
\lim_{d \to \infty} \ln p_{t,i} = -\ln \left[ 1 + (n-1) e^{-c} \right] .
\end{equation}

Hence, for a sufficiently large number of dimensions $d$, and a sufficiently large $c$, we will have $p_{t,i} \approx 1$ when $\f_t=\g_i$.

\section{Approximate $\mathcal{L}_2$ Distances in High Dimensions} \label{appendix-l2}

High dimensional vectors are known to exhibit the ``distance concentration phenomenon'' \cite{vershynin-2018,aggarwal-2001,biau-2015} where $d$-dimensional vectors with independent and identically-distributed elements tend to be concentrated on a $\mathcal{L}_p$ sphere of constant radius as $d$ grows larger. In \cite{biau-2015}, this is expressed as
\begin{equation}
\forall \epsilon > 0, \lim_{d \rightarrow \infty} P\left\{ \left| \frac{||\x||_p}{E||\x||_p} -1 \right| \ge \epsilon \right\} =0
\end{equation}
where $\x \in \mathcal{R}^d$.

It is also known that isotropic vectors tend to be orthogonal in high dimensions \cite{vershynin-2018}, i.e.,
\begin{equation}
|\mathbf{x}^T \mathbf{y}| \approx \frac{1}{\sqrt{d}}
\end{equation}
which approaches 0 for large $d$.
Acoustic neighbor embeddings are trained by modeling the ``induced'' probabilities in the embedding space as isotropic Gaussian distributions \cite{jeon-ane-2020,hinton-2002}.

Hence, for the purpose of making our analysis easier in Sections \ref{subsec:multiple-embedding} and \ref{subsec:ts-multiple-embedding}, we make the first-order approximation that the distance between any two distances will tend to be some constant $D$ in a high-dimensional space when $\x \ne \y$:
\begin{equation}
||\x - \y||^2 = \x^T \x - 2\x^T \y + \y^T \y \approx D
\end{equation}
This approximation is also a reflection of \cite{beyer-1999} where it is shown that $\L_2$ distances become more and more similar so that there is little difference between ``near'' and ``far'' in high dimensions.

\section{Maximization of Posterior for High Dimensions in Section \ref{subsec:multiple-embedding}} \label{appendix-post-high}

Defining sets $\G = \{\g_1, \cdots, \g_n \}$ and $\G' = \G \cap \{\g_a, \g_b\}^C$, we can exhaustively identify -- without using Condition \eqref{eq:maximize-posterior} -- 8 different cases for the value of $\ln p$ after accounting for symmetry
\begin{multline*} 
(i)    -4D-2\ln(n e^{-2D}) , \fone \notin \G, \ftwo \notin \G \\
\shoveleft{ (ii)   -3D-2\ln(e^{-D}+(n-1)e^{-2D}), } \\ \shoveright{ \fone=\g_a, \ftwo \notin \G }\\
\shoveleft{ (iii)  -3D-2\ln(2e^{-D}+(n-2)e^{-2D}), } \\ \shoveright{ \fone=\g_a, \ftwo \in \G' } \\
\shoveleft{ (iv)   -2D-2\ln(1+(n-1)e^{-2D}), \fone=\ftwo=\g_a } \\
\shoveleft{ (v)    -2D-2\ln(2e^{-D}+(n-2)e^{-2D}), } \\ \shoveright{ \fone=\g_a, \ftwo=\g_b } \\
\shoveleft{ (vi)   -4D-2\ln(2e^{-D}+(n-2)e^{-2D}), } \\ \shoveright{ \fone \in \G', \ftwo \in \G', \fone \ne \ftwo } \\
\shoveleft{ (vii)  -4D-2\ln(1+(n-1)e^{-2D}), \fone = \ftwo \in \G' } \\
\shoveleft{ (viii) -4D-2\ln(e^{-D}+(n-1)e^{-2D}), } \\ \shoveright{ \fone \in \G', \ftwo \notin \G } \\
\end{multline*}

Since $D>0$, it is obvious that $(iii) < (v)$ and $(vi) < (v)$.

For $(i)$, we have
\begin{eqnarray}
(i) & =   & -2\ln(n) = -2\ln(2+(n-2)) \\
    & < & -2 \ln(2+(n-2)e^{-D}) = (v) .
\end{eqnarray}

For $(ii)$, we have
\begin{eqnarray}
(ii) & =   & -2\ln(2e^{D/2} + (n-1)e^{-D/2}) \\
     & < & -2\ln(2e^{D/2} + (n-1)e^{-D/2} - e^{-D/2}) \\
     & = & (iii) < (v) .
\end{eqnarray}

For $(iv)$, we have
\begin{eqnarray}
(iv) & =   & -2\ln(e^D + e^{-D} + (n-2)e^{-D}) \\
     & < & -2\ln(2 + (n-2)e^{-D}) = (v) .
\end{eqnarray}
where we used the fact that $e^{-D}+e^{D} > 2$.

It is also obvious that $(vii) < (iv)$, so $(vii) < (v)$.

It is also obvious that $(viii) < (ii)$, so $(viii) < (v)$.

Hence, $(v)$ is the maximum among the 8 cases shown above.

\section{Maximization of Posterior for High Dimensions in Section \ref{subsec:analysis-ts-multiple-embedding}} \label{appendix-labor2}

Assuming high dimensionality, we can approximate as follows:
\begin{equation}
z_i = 
\begin{cases}
-d_i          & \textrm{if } \fone=\ftwo=\g_i \\
-D-d_i(D+1)   & \textrm{if } \fone=\g_i, \ftwo \ne \g_i \\
-2D-d_i(2D+1) & \textrm{if } \fone \ne \g_i, \ftwo \ne \g_i .
\end{cases}
\end{equation}

By Condition \eqref{eq:maximize-condition}, we have $d_a=d_b=0$, so $z_a=z_b=-D$, which means $q_a < 1/2$ and $q_b < 1/2$. Hence, $\u^{(1)}=\v_a$ and $\u^{(2)}=\v_b$ maximize $q$.

Under this constraint, we can identify 8 different cases for $\ln(q)$ = $\ln(q_a) + \ln(q_b)$ after accounting for symmetry:
\begin{multline*}\label{eq:seven_cases2}
(i)    -2D-2\ln \Big( 2 e^{-D} + e^D \sum_{j \ne a, b} h_j \Big), \\
\shoveright{ \fone \notin \G, \ftwo \notin \G } \\
\shoveleft{ (ii)   -2D-2\ln \Big( e^{-D/2} + e^{-3D/2} + e^{D/2} \sum_{j \ne a, b} h_j \Big), } \\
\shoveright{ \fone=\g_a, \ftwo \notin \G } \\
\shoveleft{ (iii)  -2D-2\ln \Big( e^{-D/2} + e^{-3D/2} + e^{D/2}m_c + e^{D/2} \hspace{-.08in} \sum_{j \ne a, b,c} \hspace{-.06in} h_j \Big), } \\
\shoveright{ \fone=\g_a, \ftwo=\g_c } \\
\shoveleft{ (iv)   -2D-2\ln \Big( 1 + e^{-2D} + \sum_{j \ne a, b} h_j \Big), } \\
\shoveright{ \fone=\ftwo=\g_a } \\
\shoveleft{ (v)    -2D-2\ln \Big( 2e^{-D}+ \sum_{j \ne a, b} h_j \Big), } \\
\shoveright{ \fone=\g_a, \ftwo=\g_b } \\
\shoveleft{ (vi)   -2D-2\ln \Big( 2e^{-D}+ e^D m_c + e^D m_d + e^D \hspace{-.08in} \sum_{j \ne a, b, c, d} \hspace{-.06in} h_j \Big), } \\
\shoveright{ \fone=\g_c, \ftwo=\g_d } \\
\shoveleft{ (vii)  -2D-2\ln \Big( 2e^{-D} + e^{D}e^{-d_c}+e^D \sum_{j \ne a, b, c} h_j \Big), } \\
\shoveright{ \fone = \ftwo = \g_c } \\
\shoveleft{ (viii) -2D-2\ln \Big( 2e^{-D} + e^{D}m_c +e^D \sum_{j \ne a, b, c} h_j \Big), } \\
\shoveright{ \fone=\g_c, \ftwo \notin \G , } \\
\end{multline*}
where $w_a, w_b, w_c, w_d$ are all distinct words in the training vocabulary, with corresponding distinct embeddings $\g_a, \g_b, \g_c, \g_d \in \G$, and we have defined
\begin{eqnarray}
     h_i & \triangleq & e^{-2D-d_i(2D+1)} \\
     m_i & \triangleq & e^{-D-d_i(D+1)} ,
\end{eqnarray}
for $i \ge 1$.

Since $D>0$, it is obvious that $(i) < (v)$.

We know $e^{D/2} + e^{-D/2} > 2$, which implies $e^{-D/2} + e^{-3D/2} > 2 e^{-D}$, and therefore
\begin{equation}
     (ii) < -2D - 2 \ln \Big( 2e^{-D} + e^{D/2} \sum_{j \ne a,b} h_j \Big) < (v) .
\end{equation}

Since $m_c > h_c$,
\begin{multline}
     (iii) < -2D - 2 \ln \Big( e^{-D/2} + e^{-3D/2} + e^{D/2} \sum_{j \ne a,b} h_j \Big) \\
     =(ii) < (v)
\end{multline}

We know $e^{D} + e^{-D} > 2$, which implies $1 + e^{-2D} > 2e^{-D}$, so
\begin{equation}
     (iv) < -2D - 2 \ln \Big( 2e^{-D} + \sum_{j \ne a,b} h_j \Big) = (v) .
\end{equation}

Since $m_c > h_c$ and $m_d > h_d$, we have
\begin{equation}
     (vi) < -2D - 2 \ln \Big( 2e^{-D} + e^D \sum_{j \ne a,b} h_j \Big) < (v) .
\end{equation}

Since $h_c < e^{-d_c}$, we have
\begin{equation}
     (vii) < -2D - 2 \ln \Big( 2e^{-D} + e^D h_c + e^D \sum_{j \ne a,b} h_j \Big) = (i) < (v) .
\end{equation}

Since $m_c > h_c$, we have
\begin{equation}
     (viii) < -2D - 2 \ln \Big( 2e^{-D} + e^D \sum_{j \ne a,b} h_j \Big) = (i) < (v) .
\end{equation}

Hence, $(v)$ is the maximum of $\ln(q)$.

\section{Word Offset and Duration Limits} \label{offset-duration-limits}

\begin{figure}[h]
     \centering
     \subfloat[]{
          \includegraphics[width=.9\linewidth]{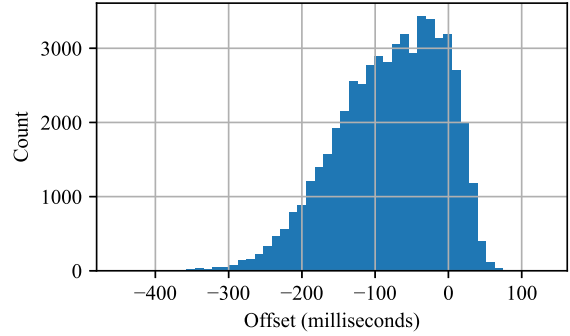}
          \label{fig:offset-hist}
     }
     \newline
     \subfloat[]{
          \includegraphics[width=.9\linewidth]{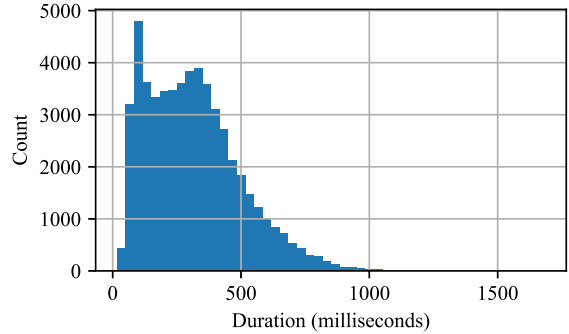}
          \label{fig:duration-hist}
     }
     \caption{Histograms of word offsets and durations produced by the proposed system for TIMIT data.}
     \label{fig:hist}
\end{figure}

In Figure \ref{fig:hist}, we show histograms of the word offsets and durations produced by the proposed system ($L=3$) system for the TIMIT data.
The minimum observed word offset was $-451$ms, and the maximum was $133$ms. 
The minimum observed word duration was $17$ms, and the maximum was $1684$ms.
All the offsets and durations are well within the $-1$s $\sim$ $+1$s and $2$s limits imposed by the hyperbolic tangent and sigmoid operators in the system, implying that the limits do not have an actual effect during evaluation.

When we removed the sigmoid and hyperbolic tangent operators and retrained the models, however, we noticed a small degradation in the word segmentation accuracy and speech recognition accuracy, implying that the operators may still help train the system more accurately.

\section{Mitigation of Insertion Errors Using Timestamps} \label{appendix-insertion}

\begin{figure}[h]
\centering
     \subfloat[]{
          \includegraphics[width=0.8\linewidth]{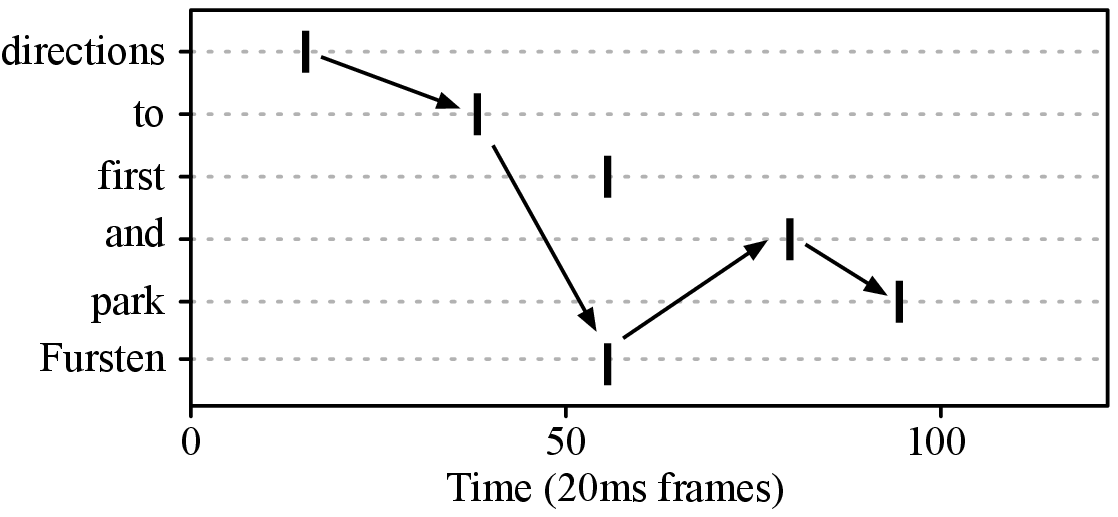}
          \label{fig:decode-baseline}
     }
     \vspace{0.2in}
     \subfloat[]{
          \includegraphics[width=0.8\linewidth]{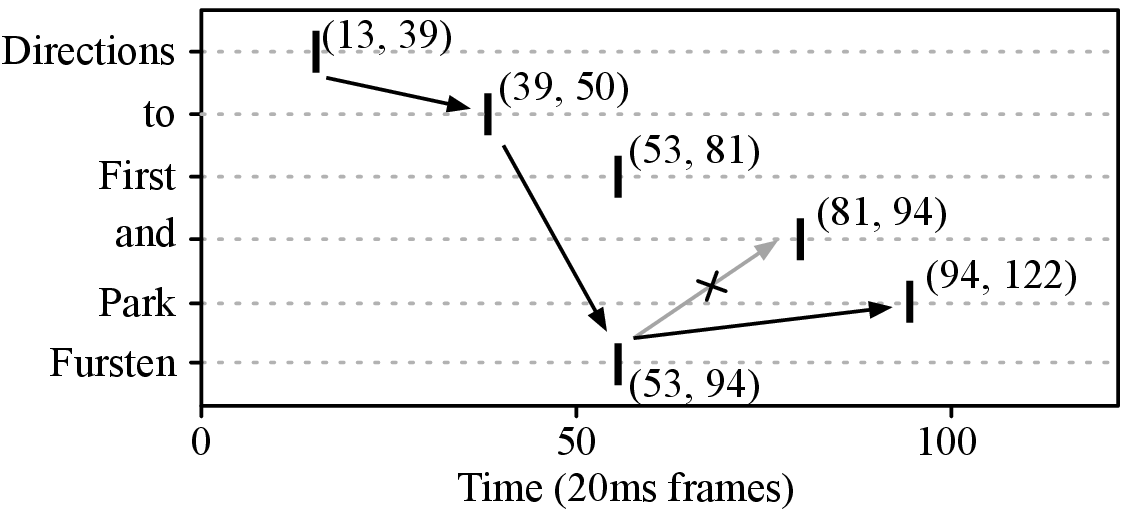}
          \label{fig:decode-timestamp}
     }
     \caption{
     Example word posterior scores on (\emph{time}$\times$\emph{word}) planes. The arrows represent example paths hypothesized by the decoder for the audio ``directions to Fursten Park.'' In (a), it is possible for the decoder to choose an erroneous path representing ``directions to Fursten and Park.'' When timestamps are present as in (b), however, the decoder can avoid the erroneous path and choose the correct one.
     }
     \label{fig:decode}
\end{figure}

Fig. \ref{fig:decode-baseline} shows a simplified set of word posterior scores produced by the embedding-matching word CTC model for the speech utterance ``directions to Fursten Park.''
Each bar ``I'' represents a spike in the score for the corresponding word and time. 
It is possible for the decoder to produce ``directions to Fursten and Park'' as the final recognition result, even though ``and'' is clearly the last part of ``Fursten'' erroneously getting recognized a second time.

In a hybrid DNN-HMM ASR system, this sort of insertion typically does not occur because every speech frame is explicitly assigned a position in a state sequence, and a speech frame that is already consumed by ``Fursten'' cannot be reused by any succeeding word. In our embedding-matching word CTC system, however, such an insertion can occur because there is no built-in systemic notion of word-frame ownership. This is exacerbated when we aggressively induce the system to hypothesize over multiple segment lengths at every point in time \cite{yen-2023} as described in Sec. \ref{subsec:multiple-embedding}.

Accurate word timestamps allow us to easily address this problem.
As shown in Figure \ref{fig:decode-timestamp}, the beam decoder can see that ``Fursten'' significantly overlaps with ``and'' and therefore refuse to follow such a path.
Hence, in the decoder for the proposed timestamped system, we add a condition that when a preceding word's end time is $t_e$ and a candidate succeeding word's start time is $t_s$, the prefix decoder refuses to join the two words if $t_e > t_s + \gamma$ where $\gamma \ge 0$ is a constant tolerance. For the results in Table \ref{tab:assistant}, we used $\gamma=200$ms.

\bibliographystyle{IEEEtran}
\bibliography{refs}

\end{document}